\documentclass[11pt]{article}

\usepackage[T1]{fontenc}
\usepackage{lmodern}
\usepackage{microtype}
\usepackage{amsmath,amssymb,amsfonts}
\usepackage{graphicx}
\usepackage{booktabs}
\usepackage{siunitx}
\usepackage{hyperref}
\usepackage{cleveref}
\usepackage{authblk}
\usepackage{enumitem}
\usepackage{natbib}

\title{Latent-Constrained Conditional VAEs for Augmenting Large-Scale Climate Ensembles}

\author[1]{Jacquelyn A. Shelton}
\author[4]{Przemys\l{}aw Polewski}
\author[2]{Alexander Robel}
\author[3]{Matthew Hoffman}
\author[3]{Stephen Price}

\affil[1]{Hong Kong Polytechnic University}
\affil[2]{Georgia Institute of Technology}
\affil[3]{Los Alamos National Laboratory}
\affil[4]{TomTom North America Inc.}

\date{Draft (tech report / preprint). December 31, 2025}

\begin{document}
\maketitle

\begin{abstract}
Large climate-model ensembles are computationally expensive; yet many downstream analyses would benefit from additional, statistically consistent realizations of spatiotemporal climate variables. 
We study a generative modeling approach for producing new realizations from a limited set of available runs by transferring structure learned across an ensemble.
Using monthly near-surface temperature time series from ten independent reanalysis realizations (ERA5), we find that a vanilla conditional variational autoencoder (CVAE) trained jointly across realizations yields a fragmented latent space that fails to generalize to unseen ensemble members. 
To address this, we introduce a \emph{latent-constrained} CVAE (LC-CVAE) that enforces cross-realization homogeneity of latent embeddings at a small set of shared geographic ``anchor'' locations. 
We then use multi-output Gaussian process regression in the latent space to predict latent coordinates at unsampled locations in a new realization, followed by decoding to generate full time series fields.
Experiments and ablations demonstrate (i) instability when training on a single realization, (ii) diminishing returns after incorporating roughly five realizations, and (iii) a trade-off between spatial coverage and reconstruction quality that is closely linked to the average neighbor distance in latent space.
\end{abstract}

\section{Introduction}
Climate model simulations are expensive, motivating methods that extract maximal information from available realizations and enable generation of additional statistically consistent samples. 
Large ensembles of smaller independent simulations allow internal variability to be characterized, but standard off-the-shelf machine learning methods often struggle to represent multiple independent realizations in a shared low-dimensional representation.
Our goal is to develop a deep generative modeling workflow that (a) captures internal variability in a low-dimensional latent space with low reconstruction error, (b) represents complex spatiotemporal data, and (c) enables generation of new realizations to reduce the cost of obtaining additional samples.

\paragraph{Contributions.}
This preprint documents progress presented at AGU Fall Meeting 2024 and makes the following contributions:
\begin{itemize}[leftmargin=*,itemsep=0.25em]
  \item Empirical evidence that jointly training a CVAE across ensemble members can yield \emph{latent fragmentation} and poor generalization to unseen realizations.
  \item A latent homogeneity constraint (LC-CVAE) that promotes alignment of latent structure across realizations using a small set of geographic anchor points.
  \item A latent-space completion strategy using multi-output Gaussian processes to predict dense latent fields for a new realization from sparse anchor samples, followed by CVAE decoding.
\end{itemize}

\section{Data and Problem Formulation}
We consider an ensemble of independent realizations of monthly reanalysis output for mean near-surface temperature from 1940--present.
Each realization provides a gridded spatiotemporal field that can be viewed as a collection of time series indexed by geographic location.
Let $x \in \mathbb{R}^{d_x}$ denote location metadata (e.g., latitude/longitude), and let $y \in \mathbb{R}^{T}$ denote the corresponding temperature time series at that location.
We assume $R$ realizations; each realization $r \in \{1,\dots,R\}$ contains a set of locations $\mathcal{X}_r$ (typically a shared grid) with observed series $\{y_r(x)\}_{x\in\mathcal{X}_r}$.

\paragraph{ERA5 monthly 2\,m temperature.}
The dataset consists of output from the ERA5 reanalysis model \cite{hersbach2023era5_monthly_singlelevels}, providing 10 independent realizations of monthly averaged mean surface temperature from 1940 to the present. Each realization comprises time series at geographic locations worldwide, capturing spatiotemporal variability influenced by climate dynamics. The data is sourced from the Copernicus Climate Change Service Climate Data Store.
The CDS product is provided on a regular latitude--longitude grid (regridded at 0.25$^\circ \times$ 0.25$^\circ$ for the reanalysis)
at monthly temporal resolution \cite{hersbach2023era5_monthly_singlelevels}.
In this work we download the period 1940--2023 and restrict to the \texttt{t2m} variable (units: K).

Future versions will detail preprocessing (e.g., normalization, spatial resolution handling) and specific subsets used in experiments.

\section{Conditional Variational Autoencoders}
A conditional VAE~\cite{sohn2015learning} models the conditional likelihood $p_\theta(y\mid x)$ using latent variables $z\in\mathbb{R}^{d_z}$ and parameter sets $\theta$ (generative) and $\phi$ (inference).
The generative process is
\begin{align}
z &\sim p_\theta(z\mid x), \\
y &\sim p_\theta(y\mid x,z),
\end{align}
with approximate posterior $q_\phi(z\mid x,y)$.
Training maximizes the evidence lower bound (ELBO):
\begin{align}
\log p_\theta(y\mid x) \ge 
\mathcal{L}_{\mathrm{CVAE}}(x,y;\theta,\phi)
&= -\mathrm{KL}\!\left(q_\phi(z\mid x,y)\,\|\,p_\theta(z\mid x)\right)
+ \mathbb{E}_{q_\phi(z\mid x,y)}\left[\log p_\theta(y\mid x,z)\right]. \label{eq:elbo}
\end{align}
In our implementation, the encoder outputs $\mu_\phi(x,y)$ and diagonal $\sigma^2_\phi(x,y)$ so that
$q_\phi(z\mid x,y)=\mathcal{N}(z;\mu_\phi(x,y), \mathrm{diag}(\sigma_\phi^2(x,y)))$.
Geographic conditioning encourages nearby coordinates to embed to similar latent codes, aiding interpretability and visualization.

\subsection{Failure mode: latent fragmentation across realizations}
When training a single CVAE jointly across all realizations into a shared latent space (we used $d_z=3$ for visualization), we observe that the latent space becomes fragmented: each realization occupies its own subspace with no discernible alignment across realizations, despite the known correspondence in geographic space.
This fragmentation prevents reconstruction and sample generation for an unseen realization. %See Fig \fig{fig:latent-fragmentation}.

\begin{figure}[t]
  \centering
  \includegraphics[width=0.8\linewidth]{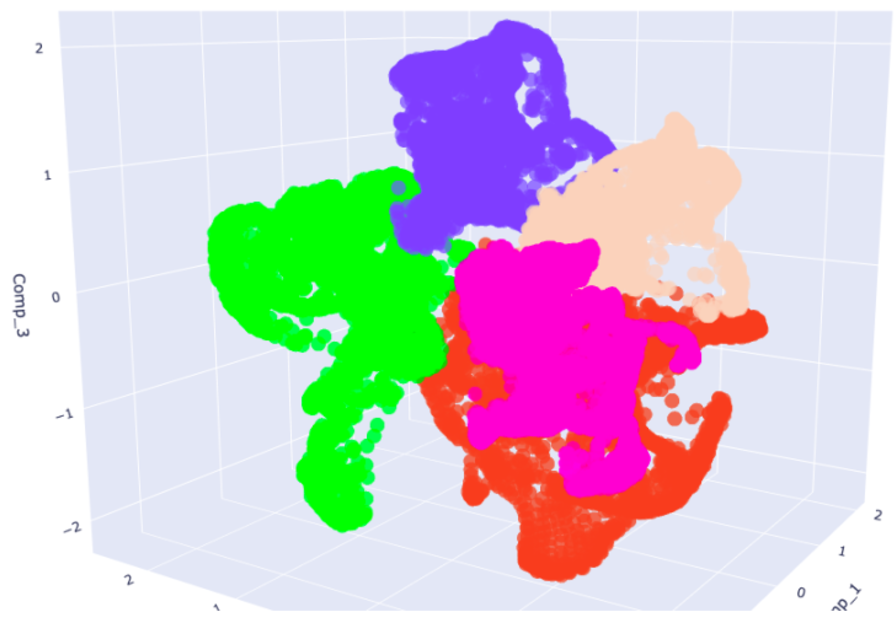}
  \caption{Latent-space fragmentation in a standard CVAE trained jointly across multiple realizations. Latent encodings tend to cluster by realization identity rather than by shared structure, degrading generalization to unseen realizations.}
  \label{fig:latent-fragmentation}
\end{figure}

\section{Latent-Constrained Conditional VAE (LC-CVAE)}
To promote a homogeneous latent structure across realizations, we train a separate CVAE per realization but add cross-realization alignment constraints at a small set of geographic \emph{anchor} locations.
Let $\mathcal{A}\subset\mathcal{X}$ denote anchor locations shared across realizations.
For each anchor $(x,y)$ we choose a ``fixed'' latent point $z^{f}_{x}$ (e.g., from a reference realization or an average embedding).
We add a penalty that enforces each realization's embedding $q_{\phi_r}(z\mid x, y_r(x))$ to remain within a maximum distance $D_{z,\max}$ of $z^{f}_{x}$.
A convenient form is
\begin{align}
\mathcal{L}_{\mathrm{LC\text{-}CVAE}}(x,y;\theta,\phi) 
= \mathcal{L}_{\mathrm{CVAE}}(x,y;\theta,\phi)
\;-\; \lambda \, \mathbb{E}_{(x,y)\sim \rho_{\mathrm{A}}}\!\left[ \left(\| \mu_\phi(x,y) - z^{f}_{x}\|_2^2 - D_{z,\max}^2 \right)_{+} \right], \label{eq:lccvae}
\end{align}
where $(\cdot)_{+}=\max(\cdot,0)$ and $\rho_{\mathrm{A}}$ is the empirical distribution over anchor samples.\footnote{The poster version uses a squared-distance constraint written in-line; \cref{eq:lccvae} is a cleaned-up equivalent with an explicit hinge. You can swap this for an exact penalty or a soft constraint if desired.}
This encourages different realizations to share a common latent geometry.

\begin{figure}[t]
  \centering
  \includegraphics[width=0.8\linewidth]{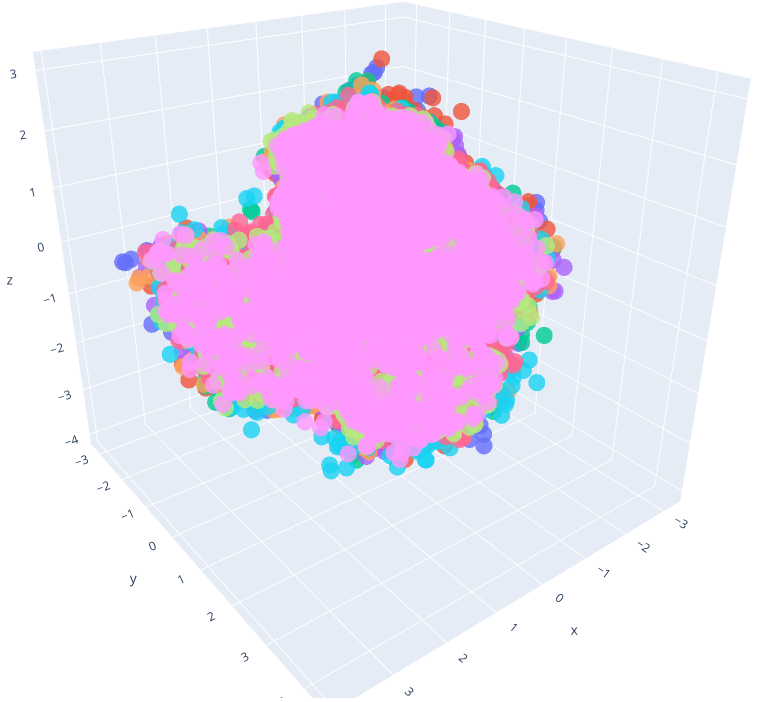}
  \caption{Effect of the latent homogeneity constraint (LC-CVAE): latents corresponding to the same conditioning context become locally aligned across realizations, mitigating the realization-driven fragmentation shown in \cref{fig:latent-fragmentation}.}
  \label{fig:latent_alignment}
\end{figure}

\section{Latent-space completion with Multi-output \\Gaussian Processes}
Even with aligned anchors, a new realization may only be observed at a small subset of locations.
We complete a dense latent field by learning a mapping from \emph{local latent neighborhood features} to latent coordinates.

\subsection{Features and regression targets}
For each realization $r$ and location $x$, let $\hat z_r(x)=\mu_{\phi_r}(x,y_r(x))$ denote the encoder mean.
Define a feature map $F_r(x)\in\mathbb{R}^{D}$ by concatenating the latent codes of the $k$ nearest neighbors of $x$ (e.g., nearest in geographic space or nearest among observed anchors) within realization $r$:
\begin{equation}
F_r(x) = \left[\hat z_r(x^{(1)}), \hat z_r(x^{(2)}), \dots, \hat z_r(x^{(k)})\right].
\end{equation}
We train a multi-output regressor $f:\mathbb{R}^{D}\rightarrow \mathbb{R}^{d_z}$ such that $f(F_r(x)) \approx \hat z_r(x)$.

\subsection{Multi-output GP model}
To predict latent coordinates for new locations in an unseen realization, we employ multi-output Gaussian process regression (GPR)\cite{vanderwilk2020framework}, a flexible nonparametric model.
We model each latent coordinate $l\in\{1,\dots,d_z\}$ as a Gaussian process:
\begin{equation}
g_l \sim \mathcal{GP}(0, k_l(\cdot,\cdot)), \qquad \hat z^{(l)} \approx g_l(F),
\end{equation}
trained via sparse variational inference. 
This provides uncertainty-aware predictions of latent coordinates at unseen locations, which we then decode with the realization-specific decoder to generate time series:
\begin{equation}
\tilde y_r(x) \sim p_{\theta_r}\!\left(y \mid x, \tilde z_r(x)\right),\quad \tilde z_r(x)=f(F_r(x)).
\end{equation}

\begin{figure}[t]
  \centering
  \includegraphics[width=0.95\linewidth]{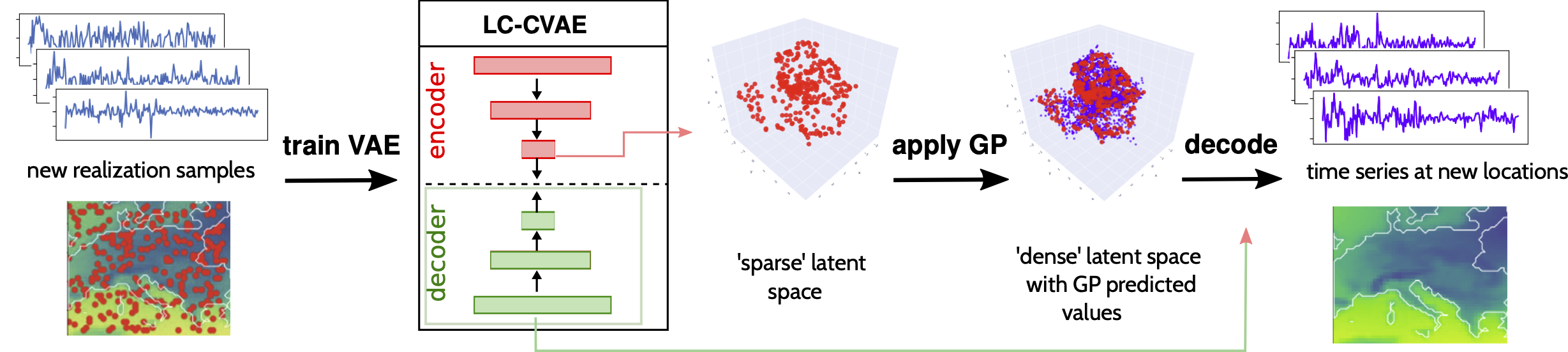}
  \caption{Schematic of latent completion for an unseen realization. Sparse latent codes inferred at observed locations are used as training targets for (multi-output) GP regression, producing dense latent codes that are decoded into a completed realization.}
  \label{fig:gp_completion}
\end{figure}

\section{Experiments}
\subsection{Setup}
We train on $R_{\mathrm{train}}$ realizations and evaluate on a held-out realization $r^\star$.
Given a coverage fraction $\alpha\in(0,1]$, we observe only $\alpha$ of the spatial locations in $r^\star$, encode anchors, predict a dense latent field via the GP, and decode to obtain reconstructed time series at all locations.

\paragraph{Metrics.}
We report mean-squared reconstruction error (MSE) over time and space. 
We also compute the average neighbor distance in latent space as a proxy for extrapolation difficulty.

\subsection{Ablations and qualitative behavior}
Experimental results suggest:
\begin{itemize}[leftmargin=*,itemsep=0.25em]
  \item \textbf{Number of realizations.} A single-realization training regime is unstable; improvements show diminishing returns after approximately five realizations.
  \item \textbf{Coverage vs quality.} Reconstruction error increases with the fraction of locations to be reconstructed, with a clear trade-off governed by the ordering of locations by neighbor distance.
  \item \textbf{Neighbor distance.} Beyond a threshold, reconstruction error correlates strongly with average neighbor distance, indicating latent-space extrapolation as a key driver.
\end{itemize}

\begin{figure}[t]
  \centering
  \includegraphics[width=0.98\linewidth]{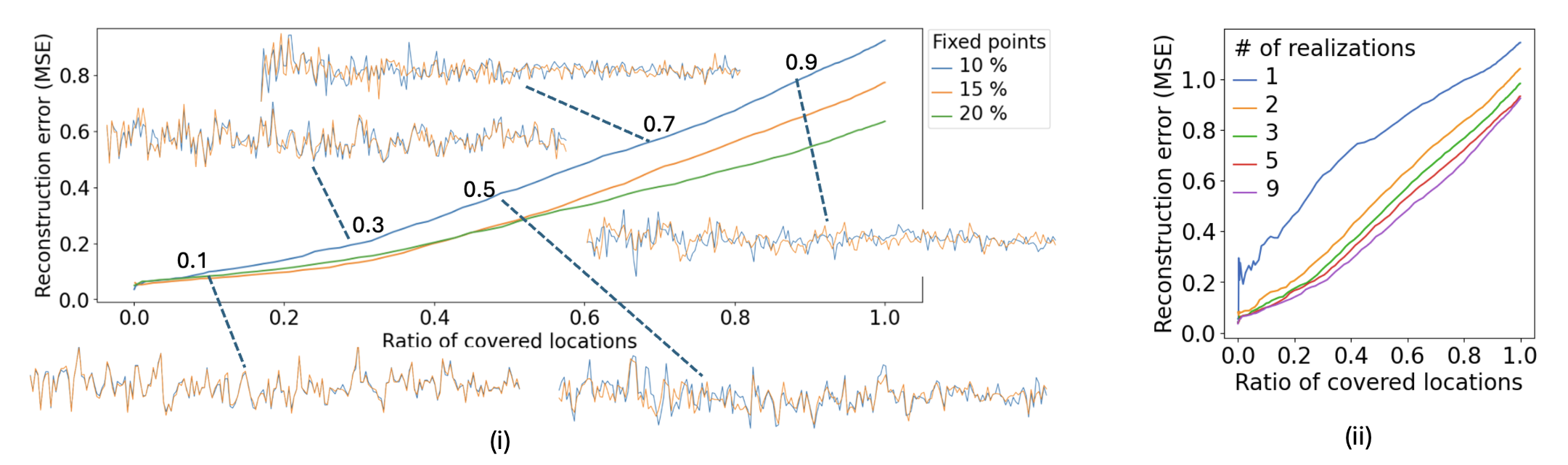}
  \caption{Ablation results from the poster study. Trends suggest instability in the low-realization regime and diminishing returns beyond a modest ensemble size.}
  \label{fig:coverage_ablation}
\end{figure}

\subsection{Sampling new realizations}
A full end-to-end run (``full pipeline'') yields new time series that visually and statistically resemble the original series at selected locations, preserving basic temporal variability and first/second-order statistics (mean and standard deviation).

\begin{figure}[t]
  \centering
  \includegraphics[width=0.98\linewidth]{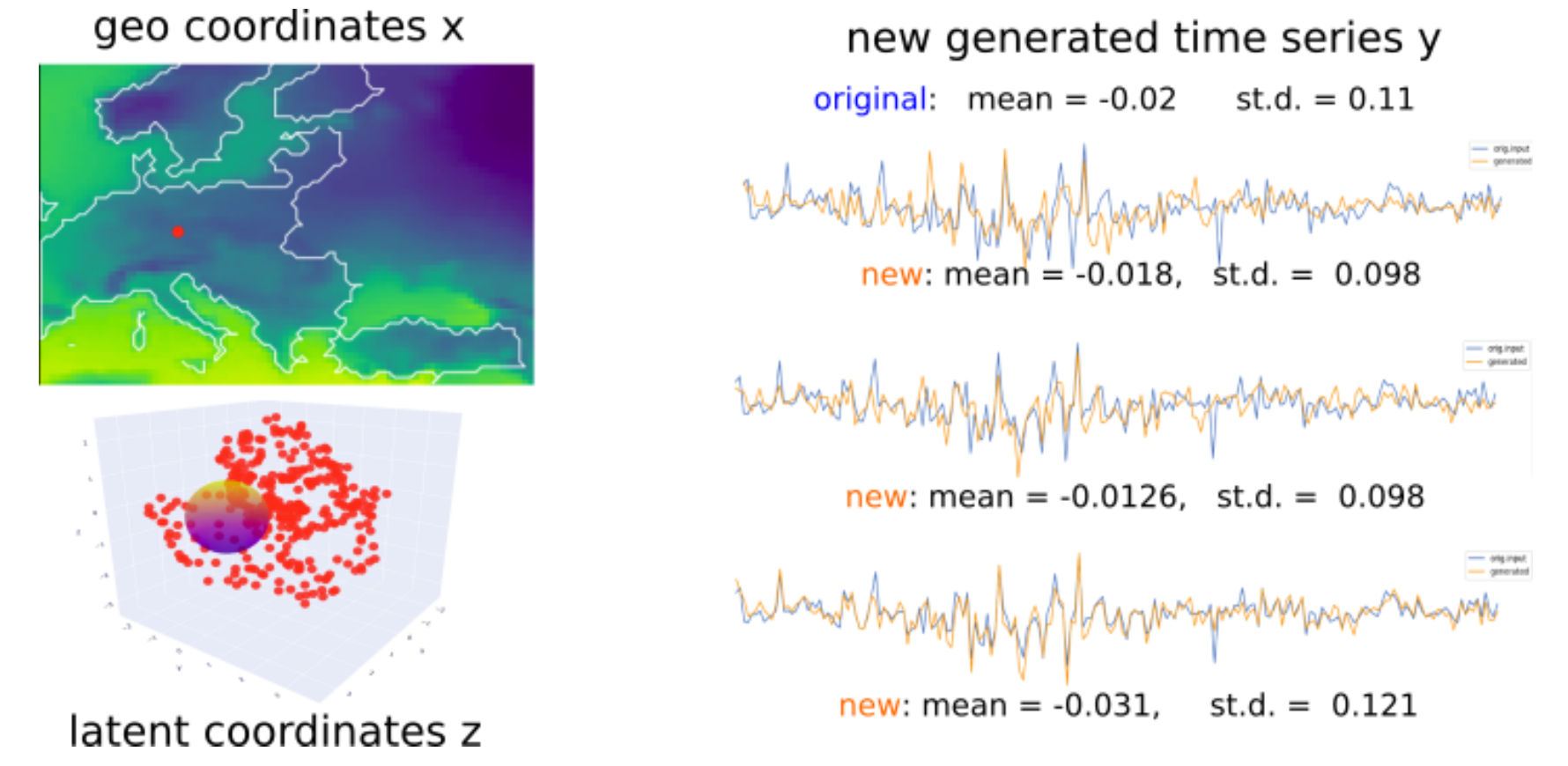}
  \caption{Qualitative examples of generation and/or completion from the poster-era pipeline. The model can generate spatially coherent fields and preserve basic temporal characteristics of the original series at selected locations.}
  \label{fig:sample_generation}
\end{figure}

\section{Discussion and Limitations}
This work should be viewed as a progress report and a baseline for more robust cross-realization generative modeling of climate ensembles.
Key limitations include: (i) reliance on a small anchor set and hyperparameters $(\lambda, D_{z,\max}, k)$, (ii) sensitivity to latent dimensionality and encoder/decoder architecture choices, and (iii) evaluation limited to reconstruction metrics and qualitative checks. 
Future work includes more rigorous distributional metrics, physically meaningful constraints, and extension to additional variables and multivariate fields.

\section{Conclusion}
We presented a workflow for generating new realizations of climate ensemble members by combining an LC-CVAE with latent-space completion via multi-output Gaussian processes.
The approach is motivated by the observed fragmentation of a jointly trained CVAE latent space and aims to transfer structure learned from multiple realizations to a new, sparsely observed realization.

%\section*{Acknowledgments}
%TODO.

\bibliographystyle{plain}
\bibliography{butt-refs}

@inproceedings{sohn2015learning,
  title        = {Learning Structured Output Representation using Deep Conditional Generative Models},
  author       = {Sohn, Kihyuk and Lee, Honglak and Yan, Xinchen},
  booktitle    = {Advances in Neural Information Processing Systems},
  volume       = {28},
  year         = {2015},
  url          = {https://proceedings.neurips.cc/paper/2015/hash/8d55a249e6baa5c06772297520da2051-Abstract.html}
}

@misc{vanderwilk2020framework,
  title        = {A Framework for Interdomain and Multioutput {G}aussian Processes},
  author       = {van der Wilk, Mark and Dutordoir, Vincent and John, ST and Artemev, Artem and Adam, Vincent and Hensman, James},
  year         = {2020},
  note         = {arXiv:2003.01115},
  url          = {https://arxiv.org/abs/2003.01115}
}

@misc{hersbach2023era5_monthly_singlelevels,
  title        = {{ERA5} monthly averaged data on single levels from 1940 to present},
  author       = {Hersbach, Hans and Bell, Bill and Berrisford, Paul and Biavati, Gionata and Hor{\'a}nyi, Andr{\'a}s and Mu{\~n}oz-Sabater, Joaqu{\'\i}n and Nicolas, Julien and Peubey, Carole and Radu, Raluca and Rozum, Iryna and Schepers, Dinand and Simmons, Adrian and Soci, Cornel and Dee, Dick and Th{\'e}paut, Jean-No{\"e}l},
  howpublished = {Copernicus Climate Change Service (C3S) Climate Data Store (CDS) [data set]},
  year         = {2023},
  doi          = {10.24381/cds.f17050d7},
  url          = {https://cds.climate.copernicus.eu/doi/10.24381/cds.f17050d7},
  note         = {Accessed: 2.2024}
}

\iffalse
\appendix
\section{Reproducibility checklist (fill in before arXiv)}
\begin{itemize}[leftmargin=*,itemsep=0.2em]
  \item Data preprocessing details (grid, anomaly removal, normalization, train/test split).
  \item CVAE architecture (encoder/decoder type, layer widths, activation, latent dim, conditional inputs).
  \item Optimization (optimizer, learning rate, batch size, epochs, early stopping).
  \item LC-CVAE hyperparameters ($\lambda$, $D_{z,\max}$, anchor fraction and selection strategy).
  \item GP details (kernel, inducing points, variational parameters, $k$ for neighbor features).
  \item Evaluation protocol (coverage fractions, location ordering, seeds, uncertainty usage).
\end{itemize}
\fi

\end{document}